\documentclass[nopreprintline, preprint, authoryear]{elsarticle}
\biboptions{longnamesfirst}
\usepackage{graphicx} % Required for inserting images
\usepackage[table,xcdraw]{xcolor}
\usepackage{booktabs}
\usepackage{bm}
\usepackage{amsmath}
\usepackage{amsfonts}
\usepackage{amsthm}
\usepackage{bbm}
\usepackage{hyperref}
\usepackage{overpic}
\usepackage{multirow}
\usepackage{float}
\usepackage[skip=2pt]{caption}
\usepackage{subcaption}
\usepackage[margin=1in]{geometry}
\usepackage{xspace} % Automatically manages spaces after macros
\definecolor{backcolour2}{rgb}{0.96,0.96,0.96}
\definecolor{backcolour}{rgb}{0.95,0.95,0.92}
\newcommand{\ps}{pointwise score\xspace}
\newcommand{\pss}{pointwise scores\xspace}

\newcommand{\nn}{\mathcal{N}}
\newcommand{\qs}{\text{QS}}
\newcommand{\is}{\text{IS}}
\newcommand{\nte}{n_{te}}

\definecolor{codegreen}{rgb}{0,0.6,0}
\definecolor{codeorange}{RGB}{250,100,0}
\definecolor{codepurple}{rgb}{0.58,0,0.82}
\definecolor{codeblue}{RGB}{50,50,250}
\definecolor{codered}{RGB}{255,40,40}

\begin{document}

\begin{frontmatter}

\title{Model selection with proper scoring rules on data sets of time series:
prefer the mean scaled score.}

\author{Giorgio Corani \corref{cor1}}
\ead{giorgio.corani@supsi.ch}

\author{Stefano Damato}
\ead{stefano.damato@supsi.ch}

\author{Dario Azzimonti}
\ead{dario.azzimonti@supsi.it}

\author{Lorenzo Zambon}
\ead{lorenzo.zambon@supsi.it}

\cortext[cor1]{Corresponding author}

\address{SUPSI, Istituto Dalle Molle di Studi sull'Intelligenza Artificiale (IDSIA),\\Lugano, Switzerland}

\begin{abstract} 
We study the problem of model selection among probabilistic forecasting models evaluated on datasets of multiple time series. The performance of a model on a single time series is quantified by the average value (\textit{score}) of a proper scoring rule over a test set, but extending model selection to data sets of  time series requires aggregating these scores. Common approaches either rely on scaling scores and averaging them (mean scaled score) or avoid scaling by using alternative statistics such as mean ranks or win rates. However, these approaches can yield conflicting conclusions.
We show that such discrepancies arise from the skewness of the distribution of the scores, which is particularly pronounced when test sets are short. The skewness can cause  non-mean criteria (e.g., mean rank, median, win rate) to select misspecified models. In contrast, the mean score is immune from this problem.
We further show that, as the size of the test sets increases, all aggregation criteria converge to the same model selection decision, mitigating these discrepancies. Our experiments on intermittent demand time series, including data from the M5 competition, highlight the importance of sufficiently large test sets; the mean scaled score 
appears to be the more reliable approach, also because empirically we found its decision to remain consistent when different scaling factors are adopted.
\end{abstract}

% \begin{keyword}
% ...
% \end{keyword}

\end{frontmatter}

\section{Introduction}
\label{scoring-rules-and-quantile-score}

\textit{Scoring rules} \citep{gneiting2007strictly,gneiting2014probabilistic} quantify how good a probabilistic forecast is.
A univariate scoring rule $\mathcal{S}$ is  a function  $\mathcal{S}(P, y)$ which assigns
a numerical score to a probability distribution $P$, given the actual observation $y \in \mathbb{R}$.
We consider  \emph{negatively oriented} scoring rules, i.e., better forecasts corresponds to lower scores.
A scoring rule is proper if it is minimized in expectation by the true distribution, i.e., if
$\mathbb{E}_{y \sim Q}\left[\mathcal{S}(Q,y)\right] \le \mathbb{E}_{y \sim Q}\left[\mathcal{S}(P,y)\right]$
for any valid forecast distributions $P, Q$. 
The most common proper scoring rules are the
continuous ranked probability score (CRPS),  the quantile score (QS), the  interval score (IS), and  the energy score (ES).

In this paper, the \textit{test set} refers to a set of
$\nte$ out-of-sample data points, related to different rolling origins and/or forecast horizons; the  \textit{\ps} is
the scoring rule evaluated on a single data point, i.e. $\mathcal{S}(P, y_i)$ for $i \in \{1, \ldots, \nte \}$, and  the \textit{score} is the
average value of the scoring rule
on the test set, i.e. $\frac{1}{\nte}\sum_{i=1}^{\nte}\mathcal{S}(P, y_i)$.
Model selection on a \textit{single} time series
is straightforward:
we select the  model with the lowest  score, assuming
$\nte$ to be large enough for the score to  approximate the expected value.

We can instead adopt different approaches for
model selection on a  data set containing $n$
time series: the most common approaches are the 
\textit{mean scaled score} and the \textit{mean rank}.
In the first case, we scale the $n$  scores to make them comparable; then we average  them and we  select the model with the \textit{lowest mean scaled score}.
While the need for scaling is established \citep{bolin_wallin_2023}, the choice of the scaling factor is not obvious; most scaling factors
\citep{hyndman_another_2006,makridakis_m5_2022,athanasopoulos2023evaluation,svetunkov2023iets} are related to the scale of the time series or to the   scale of the error of  a benchmark model. 

Generally however no scaling factor is adequate for all the non-stationarities and non-normalities which might be present in the different time series of a data set \citep{hewamalage2023forecast}.  The perceived risk is thus that the outcome of model selection might depend on the chosen scaling factor; however, to our knowledge, there is yet no  study 
on this point.

The \textit{mean rank} approach is appealing as it avoids the choice of the scaling factor. It ranks the models according to their score on the test set; then it averages the ranks over time series and selects the model with lowest mean rank.
The significance of the differences between the mean ranks is  tested with a Friedman test  \citep{demvsar2006statistical} with Nemenyi  post-hoc;  this is also  referred to as multiple comparison with the best (MCB) \citep{koning_m3_2005} in the  forecasting literature.

Yet, sometimes the mean scaled score and the  mean rank  lead to conflicting conclusions. For instance, \citet[Fig.~4]{spiliotis2021product} 
report an experiment  on the  $\sim$30,000 bottom time series of the M5 competition \citep{makridakis_m5_2022}.
On the  high quantiles, the Poisson static distribution is
competitive  according to the mean rank, but its   mean scaled score is  higher (i.e., worst) than most competitors  \citep[Tab.~1]{spiliotis2021product}. 

It remains unclear whether it is preferable to perform model selection based on the mean scaled score or the mean rank, and which criterion to trust when they lead to conflicting conclusions; in this paper
we fill this gap.
The paper is organized as follows. In Sect.~\ref{sec:pointwise-simulation}, we show that the distribution of the \pss is typically strongly skewed. In Sect.~\ref{sect:simul_model_selection}, we 
show that also the distribution of the scores, and of their differences,  is generally skewed  for common values of $\nte$. We also show that  a skewed distribution of the scores can allow  a misspecified model to have a better mean rank than the true model. The same problem is found for other non-mean statistics such as the median scaled score and the win rate.
In Sect.~\ref{sec:intermittent}, we perform model selection  on real data sets of intermittent time series, choosing between the Poisson and the negative binomial distribution.
We discuss the cases in which the mean rank and the mean scaled score  lead to different conclusions, why this happens and which criterion is arguably more reliable. We also study the sensitivity of the decisions of the mean scaled score to the  scaling factor and eventually we provide
our recommendations.

\section{The skew of the distribution of the pointwise 
scores}\label{sec:pointwise-simulation}

% intro score e dire che i pointwise score sono skewed
% introdurre mean diff and mean rank
% discutere la distribuzione della differenza dei QS, che risulta bimodale
% riusciamo a mostrare che all'aumentare di p aumenta lo skew
% quindi la distribuz delle diff è skewed, e ha implicazione sui mean rank

We now introduce the most common univariate scoring rules, 
defined to be negatively oriented. 
The quantile score  with probability $p$, $\text{QS}_{p}(y,\hat{y}_{p})$, is:
\begin{align}
\text{QS}_{p}(y,\hat{y}_{p})  = 
\begin{cases} 
2(1-p)  \big(\hat{y}_{p} - y\big) &
\text{ if } y < \hat{y}_{p} \text{ (overestimation),}
% \text{ if overestimation } (y < \hat{y}_{p})
   \\
   2(p)  \big(y - \hat{y}_{p}\big) &
\text{ if } y \ge \hat{y}_{p} \text{ (underestimation),}
   % \text{ if underestimation } (y \ge \hat{y}_{p}),
   \\
\end{cases} 
\label{eq:qscore}
\end{align}
where \(\hat{y}_{p}\) is the  quantile at level $p$ of the forecast distribution and $y$ is the observation. 

Let us discuss some properties of the  \pss of Eq.~\eqref{eq:qscore}, 
assuming $p>0.5$; the case $p<0.5$ is symmetric. 
They aee skewed since they  are lower-bounded by zero and without upper bound.
Moreover, the skew increases with $p$ as we explain in the following. The \pss are drawn from a  mixture with two components.
The overestimation component  yields many  small \pss, while  the underestimation component yields few large \pss.
Given an underestimation and an overestimation error of the same size,
underestimation results in a \pss which is $k=\frac{1-p}{p}$  times larger than overestimation.
For $p=\{0.8, 0.9, 0.99\}$, we have $k=\{4,9,99\}$.
We also notice that, for  well calibrated forecasts, the underestimation errors are $k$ times rarer than the overestimation ones.

The \textit{interval score} ($\text{IS}_{p}$) 
\citep{gneiting2007strictly}
for a prediction interval with coverage $p$ is:
\begin{align}
    \text{IS}_{p}(l,u,y) & 
= (u-l) + \frac{2}{1-p}(l-y) \mathbbm{1}\{ y < l\} +  \frac{2}{1-p}(y-u) \mathbbm{1}\{ y > u\}  \\
& = \frac{2}{1-p}  \left[ 
\text{QS}_{\frac{1-p}{2}}(y,l) + \text{QS}_{\frac{1+p}{2}}(y,u)
 \right]
 \label{eq:iscore}
\end{align}
where $l$ and $u$ are the lower and upper endpoints,
and $\mathbbm{1}$ is the indicator function, which is equal to 1 if the argument is true, and 0 otherwise.
We assume  a symmetric prediction interval, for which  $l$ and $u$ are the quantiles at level $\frac{1-p}{2}$ and $\frac{1+p}{2}$.
Given the relation between $\is_q$ and $\qs_p$, in general also the skew of the distribution  of the \pss of $\is_q$ increases with $p$.

The Continuous Ranked Probability Score (CRPS) scores the entire predictive distribution and
corresponds to the average of the quantile scores over all values of  $p$:
\begin{align}
    \text{CRPS}(F,y)= \int (F(x)- \mathbbm{1}{\{y<x\}})^2 dx =  2 \int \text{QS}_{p}(y,\hat{y}_{p}) dp,    
    \label{eq:crps}
\end{align}
where $F$ is the cumulative distribution function of the forecast distribution.
More details on the relation between QS, IS and CRPS can be found in \citet{fakoor2023flexible,Tibshirani2023ForecastScoring}.

\renewcommand{\arraystretch}{1.3}
\begin{table}[!ht]
\centering
\begin{tabular}{l c c c c c}
\toprule
 $p$ & 0.5 & 0.8 & 0.9 & 0.95 & 0.99 \\
\midrule
\rowcolor{backcolour}
$\is_p$ & 1.9 & 3.7 & 5.5 & 8.2 & 19.5 \\
$\qs_p$ & 1.0 & 2.2 & 4.1 & 6.9 & 18.3 \\
\rowcolor{backcolour}
CRPS & \multicolumn{5}{c}{1.9} \\
\bottomrule
\end{tabular}
\caption{Skewness of the distribution of the \pss of different scoring rules on  10$^5$ data points.
The $N(0,1)$ is used both as true distribution and as forecasting distribution. 
}
\label{tab:skew}
\end{table}

% \begin{table}[!h]
% \centering
% \begin{tabular}{lrr}
% \toprule
% Score & Skew $\mathcal{N}(0,1)$ & Skew of the diff\\
% \midrule
% $\qs_{0.8}$ & 2.2 & 1.4\\
% \rowcolor{backcolour}
% $\qs_{0.9}$ & 4.1 & 2.4\\
% $\qs_{0.95}$ & 6.9 & 3.5\\
% \rowcolor{backcolour}
% $\qs_{0.99}$ & 18.3 & 7.7\\
% $\is_{0.8}$ & 3.7 & 1.2\\
% \rowcolor{backcolour}
% $\is_{0.9}$ & 5.5 & 2.2\\
% $\is_{0.95}$ & 8.2 & 3.3\\
% \rowcolor{backcolour}
% $\is_{0.99}$ & 19.5 & 7.2\\
% CRPS & 1.9 & 0.7\\
% \bottomrule
% \end{tabular}
% \end{table}

We now quantify the skewness of the \ps by 
drawing $10^5$ actual values $y$ from the standard Gaussian $\nn(0,1)$ and scoring the $\nn(0,1)$ on each of them.
Thus, we use the $\nn(0,1)$  both as \emph{true} distribution and  forecasting distribution.
We then measure the skewness of the \pss  as 
$M_3 \,/\, M_2^{3/2}$ where $M_j = \sum_i^n (x_i - \bar{x})^j \,/\, n$.
The skewness is a dimensionless measure, which we
classify  as \textit{strong} and \textit{extreme}
when its absolute value respectively  exceeds 1 and 5 \citep{bulmer1979principles}.
In our experiment (Tab.~\ref{tab:skew}), the \pss of CRPS, 
$\is_{p}$ and $\qs_{p}$ with $p=0.5$ or
$p=0.8$ are \textit{strongly} skewed;
they become instead  \textit{extremely} skewed  as $p$ increases to $p=0.95$ and $p=0.99$.
At the same level $p$,  $\text{IS}_{p}$ is more skewed than
 $\text{QS}_{p}$, since
$\text{IS}_{p} = \text{QS}_{\frac{1-p}{2}}$ + $\text{QS}_{\frac{1+p}{2}}$ (Eq \ref{eq:iscore}). \\

\section{Model selection on a data set of $n$ time series}\label{sect:simul_model_selection}
In the following simulation we perform model selection between two static distributions:
the \textit{true} model  $\nn(0, 1)$ and the \textit{misspecified} model $\nn(0, 0.85^2)$. Notice that for any $p>0.5$ the misspecified model has a  smaller  $\hat{y}_p$ than the  true model.  
We sample $n=10^5$ test sets  of length $\nte$
from the true model; on  each test set we measure the average  $\qs_{0.9}$ and $\qs_{0.99}$
of both models. We consider $\nte \in \{4, 28, 100\}$.
In this experiment no scaling is necessary, so the \textit{mean scaled score} is simply the mean  of the $n$ scores.
To implement the \textit{mean rank}, on each time series we assign rank 1 and 2 to respectively the model with lower and higher score (there are no ties); we   average the ranks over time series and we select the model with the lowest mean rank.

\begin{table}[!h]
\centering
\begin{tabular}{clcccccc}
\toprule
\multicolumn{2}{c}{ } & \multicolumn{2}{c}{\textbf{CRPS}} & \multicolumn{2}{c}{\textbf{QS}$\mathbf{_{0.9}}$} & \multicolumn{2}{c}{\textbf{QS}$\mathbf{_{0.99}}$} \\
 &  & Mean score & Mean rank & Mean score & Mean rank & Mean score & Mean rank\\
\cmidrule(lr){3-4} \cmidrule(lr){5-6} \cmidrule(lr){7-8}
 & \textit{True} & \cellcolor{backcolour}\textcolor{codegreen}{\textbf{0.56}} & \cellcolor{backcolour}\textcolor{codegreen}{\textbf{1.45}} & \cellcolor{backcolour}\textcolor{codegreen}{\textbf{0.35}} & 1.59 & \cellcolor{backcolour}\textcolor{codegreen}{\textbf{0.05}} & 1.91\\
\multirow{-2}{*}{\centering\arraybackslash $\nte = 4$} & \textit{Misspec.} & 0.57 & 1.55 & 0.36 & \cellcolor{backcolour}\textcolor{red}{\textbf{1.41}} & 0.06 & \cellcolor{backcolour}\textcolor{red}{\textbf{1.09}}\\
\addlinespace[0.5em]
 & \textit{True} & \cellcolor{backcolour}\textcolor{codegreen}{\textbf{0.56}} & \cellcolor{backcolour}\textcolor{codegreen}{\textbf{1.31}} & \cellcolor{backcolour}\textcolor{codegreen}{\textbf{0.35}} & \cellcolor{backcolour}\textcolor{codegreen}{\textbf{1.38}} & \cellcolor{backcolour}\textcolor{codegreen}{\textbf{0.05}} & 1.58\\
\multirow{-2}{*}{\centering\arraybackslash $\nte = 28$} & \textit{Misspec.} & 0.57 & 1.69 & 0.36 & 1.62 & 0.06 & \cellcolor{backcolour}\textcolor{red}{\textbf{1.42}}\\
\addlinespace[0.5em]
 & \textit{True} & \cellcolor{backcolour}\textcolor{codegreen}{\textbf{0.56}} & \cellcolor{backcolour}\textcolor{codegreen}{\textbf{1.16}} & \cellcolor{backcolour}\textcolor{codegreen}{\textbf{0.35}} & \cellcolor{backcolour}\textcolor{codegreen}{\textbf{1.29}} & \cellcolor{backcolour}\textcolor{codegreen}{\textbf{0.05}} & \cellcolor{backcolour}\textcolor{codegreen}{\textbf{1.37}}\\
\multirow{-2}{*}{\centering\arraybackslash $\nte = 100$} & \textit{Misspec.} & 0.57 & 1.84 & 0.36 & 1.71 & 0.06 & 1.63\\
\bottomrule
\end{tabular}

\caption{
Model selection over $10^5$ simulations.
For the selected models we highlight the background;
we use green when the true model is selected, and red otherwise.
}
\label{tab:QS}
\end{table}

As reported in Tab.\ref{tab:QS},
on the CRPS both criteria select the true model.
Yet,  the misspecified model  has lower
 mean rank than the true model
when assessing $\qs_{0.9}$ or $\qs_{0.99}$ 
with small $\nte$
How is this possible?
Since the overestimation error (which are the common ones) are penalized proportionally to  $\hat{y}_p-y$, frequently
the misspecified model has lower \ps than the true one.
If the elicited quantile is high and/or $\nte$ is small, 
many test sets  contain no underestimation errors; thus most scores are optimistically biased compared to the actual  expected value of the scoring rule. 
By ranking  independently the models on each test set the mean rank   overlooks the importance the underestimation errors, which are present only on few test sets.
The problem progressively disappears with  larger $\nte$, which makes the distribution of the score more symmetric  around the expected value.
For instance, dealing with $\qs_{0.99}$, the mean rank selects the
misspecified model for $\nte=4$ and $\nte=28$; eventually, it selects the
true model  for $\nte=100$.
Because of the central limit theorem
the distribution of the difference of the scores
becomes symmetric as $\nte$ increases. 
However the higher the skewness of the population, the slower the convergence of the sample means   to a normal distribution  \citep[Chap. 3.4]{Durrett2019}.
For the common  values of $\nte$, the distribution of the score might still be skewed; we show that this can lead the mean rank and the mean scaled score to conflicting decisions.
In our experiment, the CRPS does not appear to be problematic, since  its \pss and thus also its (differences of) scores are less skewed compared to $\qs_{0.9}$
and $\qs_{0.99}$.

\section*{Discussion}
Let us denote by $s_i^A$ and $s_i^B$ the score of models  A and B
on time series $i$, and as $d_i$ their difference, i.e., 
$d_i = s_i^A-s_i^B$. 
Let us assume for simplicity that there are no  ties  and no scaling factors. 
The mean scaled score  selects A if  $\frac{1}{n}\sum_i s_i^A < \frac{1}{n}\sum_i s_i^B$,
which implies $\frac{1}{n}\sum_id_i <0$, i.e.,  that the \textit{mean difference} is negative.
The rank of model A on time series $i$ is
$r^A_i=\mathbbm{1}(d_i>0)+1$,
where $\mathbbm{1}$ is the indicator function.
The mean rank criterion selects A if $\bar{r}^A=\frac{1}{n}\sum_i r^A_i<1.5$, or equivalently, if $\frac{1}{n}\sum_i \mathbbm{1}(d_i>0) < 0.5$.
This means that more than half of the times $d_i<0$, i.e.,  the \textit{median difference} is negative.
The mean scaled score and the mean rank are thus  estimators of different 
parameters, which can lead to the different decisions. 
% \Ste{Vogliamo togliere il commento alla distribuzione di $d_i$? Ho ricopiato il seguente brano: \\ Crucially, 
% the median difference is equal to the mean difference only if the $d_i$'s are symmetrically distributed, which happens if $\nte$ is large enough.
% %the median difference corresponds to the difference of the expected values only if the $d_i$'s are symmetrically distributed, i.e., if $\nte$ is large enough.
%A larger $\nte$ is needed if  the distribution of the \ps is more skewed; for instance
% %$\nte$=28 is enough for the mean rank to select the true model with $\qs_{0.9}$, but not with $\qs_{0.99}$.}\Gio{l'ho tolto perchè non mi sembrava così utile.}

\begin{figure}[h!]
    \centering
    \includegraphics[width=0.8\linewidth]{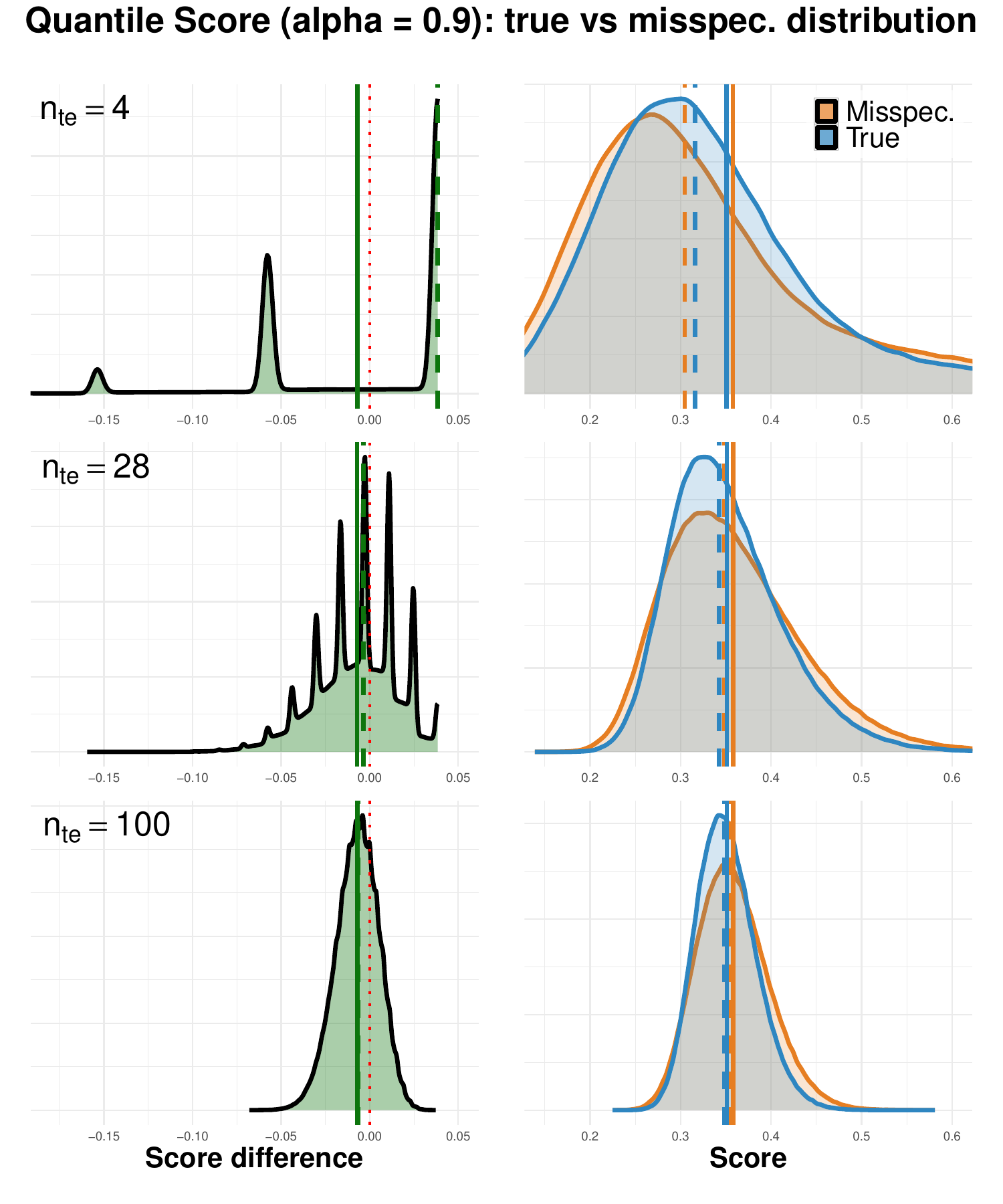}
    \caption{
    % \Lor{Left: ...., Right: ....}\\
    Distribution of the average QS over $h=4$ (top) and $h=28$ (bottom) steps ahead.
    The true distribution $\nn(0, 1)$ is compared against a misspecified distribution $\nn(0, 0.85^2)$.
    While the mean of the QS (solid line) is consistently lower for the true distribution, the median (dashed line) is higher when the test set is short. 
    This corresponds to a mean rank exceeding 1.5, leading to wrong model selection. 
    }
    \label{fig:QL_comparison}
\end{figure}

We now re-analyze the simulation of Sect.~\ref{sect:simul_model_selection} focusing on $\qs_{0.9}$.
In Fig.~\ref{fig:QL_comparison}, the right panel shows the distribution of the scores of the true and misspecified models, while the left panel shows the distribution of their difference (true $-$ misspecified), for different values of $\nte$.
For $\nte = 4$ (top), the mean difference is negative, but the median is positive: the mean score correctly selects the true model, while the mean rank selects the misspecified model (Table~\ref{tab:QS}). 
Indeed, the distribution of the differences $d_i$ is strongly skewed and multimodal.

The right peak corresponds to the simulations where no actual value $y$ exceeds  the predicted quantiles; in these cases, the score of the misspecified model is lower, because its overestimation errors are smaller. Since the test set is short, this happens frequently,  causing the median difference to be positive.
Since the test set is short, this happens on many test sets; thus,
the median difference is positive.
The other density peaks correspond to cases where, among the $\nte=4$ actual values, one or more fall in the right tail of the distribution.
These cases correspond to negative score differences, as the underestimation errors of the true model are smaller, and move the mean of the distribution of the score differences below zero.
As $\nte$ grows, the median is pushed towards the mean: 
the central limit theorem makes the distribution smoother and more symmetric, as the peaks become more numerous and closer to each other.
Eventually, for $\nte=100$, the distribution is practically Gaussian.

The mean rank is not a reliable criterion when $\nte$ is not sufficiently large. 
Proper scoring rules are minimized by the true distribution only in expectation: 
the mean rank correctly identifies the best model only if the score, obtained as the mean of the $\nte$ \pss, is an accurate approximation of this expected value. 
For the quantile score, this requires having observed an adequate number of underestimation losses, which become more rare for higher quantiles.
Determining exactly when $\nte$ is ``large enough'' remains difficult, as it depends on both the chosen scoring rule and the true distribution of the actual values. 
Therefore, we offer a general recommendation: practitioners should exercise caution when relying on mean ranks, especially when evaluating the quantile score at high quantile levels. 
In scenarios where model selection criteria yield conflicting results, the mean scaled score should generally be regarded as the more reliable metric.

The same recommendations apply to other model selection criteria.
% which are  based on  estimating the expected value of the score.
For instance, the \textit{win rate}, used  by \citet{shchur2025fev}  to compare foundation forecasting models,
is the fraction of time series on which each model ranks first.
When comparing two models,  it is equivalent to the mean rank and  thus suffers from the same limitations. 
With more than two models, the win rate is no longer equivalent to the mean rank, but it remains a function of the ranks alone and is therefore subject to the same issues.
Since it estimates how often a model achieves the lowest score rather than estimating the expected score,
it can disagree with the mean scaled score whenever $\nte$ is too small.

The \textit{median scaled score} can also be problematic.
In the top right of Fig.\ref{fig:QL_comparison}, we show that the true model has lower mean but higher median scaled score than the misspecified model.
As $\nte$ grows, the median scores move closer the mean scores, eventually yielding the same model selection.
We refer to \citet[Tabs.~2–4, E.4]{svetunkov2023iets} for an 
empirical example on the M5 bottom-level series ($\nte=28$) where mean and median scaled scores lead to conflicting conclusions at the 99th quantile.
For example, the negative binomial model is among the best models according to the mean, but among the worst according to the median. 

% \Lor{se citassimo un nostro paper come esempio?} \Ste{secondo me ci sta}\\
Another family of criteria is based on \textit{relative scores}, obtained by dividing the score of each model by that of a reference model and then aggregating across series, typically via a geometric mean.
The \textit{skill score}, defined as the relative improvement over the reference model, is essentially equivalent.
These criteria have the practical advantage of not requiring scaling factors, since the ratios are already dimensionless.
However, since they combine the scores nonlinearly, they do not estimate the expected values of the scores;
they are thus reliable only when $\nte$ is large enough for the scores to closely approximate their expectations.
\citet{wheatcroft2019interpreting} reaches a similar conclusion, showing that the skill score is biased due to the ratio in its definition.

Table~\ref{tab:median_rel_score} in \ref{app:median_scores} illustrates these points on the same simulation as Sect.~\ref{sect:simul_model_selection}, 
reporting both the median and the relative score.
We show that both criteria can yield wrong model selection, especially for small $\nte$ and for the quantile score with high quantile levels.

\section{Empirical model selection between the Poisson and the negative binomial distribution}
\label{sec:intermittent}
We now compare the mean scaled score and  the mean rank 
in a model selection experiment  on different data sets of intermittent demand. 
We also check the sensitivity of the decision of the mean scaled score on the  scaling factor. 
Our candidate models
are the Poisson and the negative binomial  distribution.
The negative binomial  is generally recognized \citep{syntetos2012demand, kolassa_commentary_2022}
as a more suitable model  than the Poisson for intermittent demand since,  thanks to overdispersion,  it can better estimate the probability of demand spikes \citep[Chap. 5]{boylan_intermittent_2021}.

\paragraph*{Experimental design} 
We consider the data sets listed in Tab.~\ref{tab:datasets}; as scoring rules we  evaluate CRPS and several quantile scores ($\qs_{0.75}$, $\qs_{0.835}$, $\qs_{0.975}$, $\qs_{0.995}$). 
We assess the impact of $\nte$   by varying it  from 1 up to the full available test set length (28 for M5; 6 for Auto and 12 for RAF).

\begin{table}[!ht] 
\centering
\begin{tabular}{lrrrrrrr}
\toprule
Dataset & \# of ts & freq. & T & max $\nte$ &  avg. ADI  \\
\midrule
\rowcolor{backcolour}
M5  (bottom only)      & 30490 & D & 1941 & 28 & 6.2 \\
%UCI       & 1191  & D & 76   & 14 & 10.1 \\
Auto      & 3000  & M & 18   & 6  & 1.3  \\
%Carparts  & 2493  & M & 45   & 6  & 5.6  \\
\rowcolor{backcolour}
RAF       & 5000  & M & 72   & 12 & 9.8  \\
\bottomrule
\end{tabular}
\caption{The  considered datasets; ``D" stands for daily and ``M" for monthly; $T$ 
denotes the length of the training and test set. Avg. ADI is the average inter-demand interval, averaged  over all time series of the data set. We took the M5 data set from the \href{https://www.kaggle.com/competitions/m5-forecasting-accuracy/data}{Kaggle page of the competition} and the Auto and RAF data sets from \citet{fableintermittent}.}
\label{tab:datasets}
\end{table}

\begin{table}[!ht]
    \centering
    \begin{tabular}{cc}
    \toprule
        Scaling factor & Formula  \\
        \midrule
        \rowcolor{backcolour}
         MAE naive & $\frac{1}{T-1}\sum_{i=1}^{T-1} |y_{i+1} - y_i|$\\
         in-sample mean & $\frac1T \sum_{i=1}^{T} y_i$ \\
         \rowcolor{backcolour}
         in-sample score of ED & $\frac{1}{T} \sum_{i=1}^T \mathcal{S}(y_i, \text{ED})$  \\
         \bottomrule
    \end{tabular}
    \caption{The considered scaling factors. We denote the training set as $y_1, \dots, y_T$. The in-sample score of ED is the mean score achieved on the training set by the empirical distribution of the data. }
    \label{tab:scaling}
\end{table}

We consider the scaling factors of Tab.\ref{tab:scaling}.
The \textit{MAE naive} was  introduced as scaling factor of the MASE \citep{hyndman_another_2006}.
It was used to scale scoring rules in the M5 competition \citep{makridakis2022m5}; its main shortcoming is that it might inflate the influence of the most intermittent series.
The \textit{in-sample mean
} is an appropriate scaling factor for stationary time series but it might be inadequate for data sets containing a mixture of stationary and trending series \citep{athanasopoulos2023evaluation,hewamalage2023forecast}.
It has been  used for instance by \citet{svetunkov2023iets}.

The \textit{in-sample score of ED} 
scaling factor is constituted by the average value of the scoring rule  on the training data, achieved by a reference model.
As in \citet{damato2025forecasting},
our reference model is
the empirical distribution (ED) of the data, which 
is  a sound probabilistic baseline for intermittent time series \citep{spiliotis2021product, kolassa_commentary_2022, long_scalable_2025}. When the scoring rule is QS$_p$, the scaling factor corresponds to the in-sample score of the empirical quantiles.

\begin{figure}[!ht]
    \centering
    \includegraphics[width=0.99\linewidth]{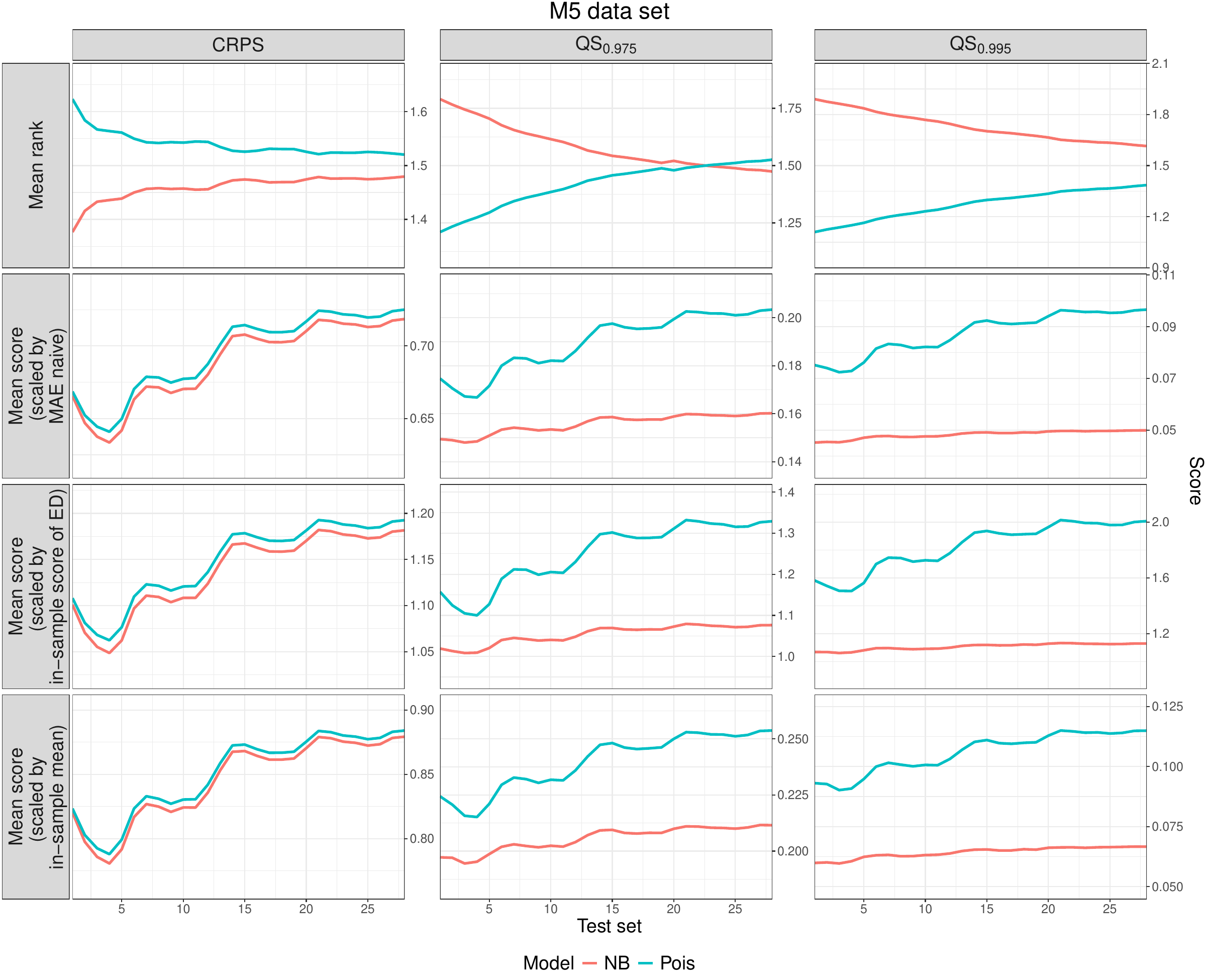}
    \caption{Comparison of the Poisson and the negative binomial distribution on
on the M5 data set, considering different scoring rules, scaling factors and test set lengths.
% \Gio{ottime. serve ancora piccoli ritocchi: Test set diventa $\nte$, font degli assi più grandi,
% in legenda i nomi estesi delle distribuzioni}
}   
    \label{fig:M5_experiment}
\end{figure}

We now analyze the results on the bottom time series of the M5 dataset. In Fig.~\ref{fig:M5_experiment} each column corresponds to a different scoring rule and each row to a different model selection criterion, namely the mean rank or the mean scaled score with various scaling factors. Depending on the row, we select the model with either the lower rank or the lower mean scaled score. Each panel shows how the mean ranks and mean scaled scores vary with $\nte$.
To improve readability, Fig.~\ref{fig:M5_experiment} only shows CRPS, $\qs_{0.975}$ and $\qs_{0.995}$. In Fig~\ref{fig:M5_experiment_full} we show the full results. 

On  CRPS (Fig.~\ref{fig:M5_experiment}, col.1) mean rank and mean scaled score with any scaling factor  select the negative binomial at any value of $\nte$. 
They also  agree also on $\qs_{0.835}$ and $\qs_{0.75}$ (only shown in Fig~\ref{fig:M5_experiment_full}), 
this time selecting the Poisson. Indeed, different models can be preferable for different scoring rules.

For $\qs_{0.975}$, the decision of the mean rank criterion is sensitive on $\nte$
(Fig.~\ref{fig:M5_experiment}, col.2, row 1):
it selects the  Poisson if provided with only a short test set and instead  the negative binomial
if provided with a long enough test set.
Indeed, the mean rank of the negative binomial and the Poisson 
has respectively a strong downward and upward trend with $\nte$.
Instead, the mean scaled score selects the negative binomial model for all $\nte$ with any scaling factor (Fig.~\ref{fig:M5_experiment}, col.2, rows 2–4).
We assume the negative binomial to be the correct decision, being both chosen when larger test sets are available and supported by the literature.

The behavior of the mean rank  can be explained as the simulation of Sect.~\ref{sect:simul_model_selection}: given  a negative binomial and a Poisson distribution with the same mean, the Poisson has a lower $p$-th quantile for $p > 0.5$ (although the two quantile forecasts may coincide in some cases due to discreteness). Dealing with high quantile levels and small $\nte$, many test set  contain no underestimation error; thus the Poisson  
might has lower score than the negative binomial on them, obtaining eventually a better mean rank. 
As $\nte$ increases, however, the decisions of  the mean rank converges to those of the mean scaled score.
Thus the mean rank can  require substantially larger $\nte$ than the mean scaled score  to stabilize its decision when the distribution of the difference of the scores is strongly skewed,
as it is typically the case of scoring rules with high levels of $p$.
% The mean scaled score instead correctly accounts for the rare underestimation errors even with small $\nte$, by averaging over all data points of all test sets. 

% The estimated reduction in mean scaled score achieved by the negative binomial compared to the Poisson varies between 12\% and 21\% depending on $\nte$; notably, 
% given the same $\nte$, 
% the estimated percentage reduction of the score is broadly consistent across the different scaling factors (this consideration applies to practically all scores and data sets considered in our experiments).

On $\qs_{0.995}$ the issue is even more apparent.
Also in this case the mean ranks of the two models show  a strong
trends with $\nte$: upward  for the Poisson,  downward for the negative binomial. 
We assume that, if we could further increase $\nte$,  the mean rank 
of the negative binomial would eventually become lower than that of the Poisson.
However, the available test set is too short too observe this; as a result,
the mean rank criterion selects the Poisson model for all $\nte$ (Fig.~\ref{fig:M5_experiment}, row 1, col.3). 
Instead, the mean scaled score selects the negative binomial 
for all values of $\nte$, with any scaling factor.
This conflicting conclusions is reported also by \citep[][Tab.~1 and Fig.~4]{spiliotis2021product},
who considered $\nte=28$ and the \textit{mae naive} scaling factor, and whose experiments included the  negative binomial and the Poisson among many competing models.  Indeed, they acknowledge the conflicting conclusions achieved by the mean rank and the mean scaled score.
We extended their analysis by studying the sensitivity to $\nte$ and to the scaling factor.

We report in Fig.~\ref{fig:Auto_experiment_full} and Fig.~\ref{fig:RAF_experiment_full} the results for the Auto and RAF data sets, which have smaller test sets (6 and 12 respectively). 
On both data sets we observe the same patterns.
Mean rank and mean scaled score with any scaling factor take the same decisions on CRPS, $\qs_{0.75}$ and $\qs_{0.835}$ (only  CRPS is shown in the figures). 
On $\qs_{0.975}$ and $\qs_{0.995}$, on both data sets, the mean rank
criterion selects the Poisson distribution at any $\nte$; yet, the negative binomial has a strong downward trend with $\nte$, similarly to the previous cases.
We assume that, with a larger test set, the mean rank criterion would eventually switch its decision in favor of the negative binomial, on both quantiles and on both data sets.
Instead, on both quantiles and on both data sets, the mean scaled score selects the negative binomial at any $\nte$ and with any scaling factor. 

RAF is the  dataset with highest ADI (Tab. \ref{tab:datasets}) and thus the highest percentage of zeros. In this case,  assuming the  variance to be  equal to the mean is  especially limiting .
Indeed, the mean scaled scored selects the negative binomial 
for any quantile, value of $\nte$ and scaling factor. Instead, the mean ranks  selects the negative binomial distribution on $\qs_{0.975}$ and $\qs_{0.995}$.  In both cases, there is a strong downward trend of the  mean rank of the negative binomial model, which suggest that decision of the mean rank could eventually switch to the negative binomial, if a longer test was available.

In general, the decision of the mean scaled score is  consistent at any $\nte$ and with any scaling factor. 
The estimated percentage difference between models is also broadly consistent across scaling factors, although it has some sensitivity on $\nte$ (in our experiments, the percentage improvements in mean scaled provided by the negative binomial over the Poisson tends to increase with $\nte$).
% the estimate of the improvement provided by the negative binomial over the Poisson stabilizes when the test set is large enough. On quantile 0.995, it starts at about 32\%
% at $\nte = 1$ and stabilizes at about 45\% for $\nte$ of about 15 or more.

\section{Conclusions}\label{conclusions-and-recommendations}

The distribution of the scores (and of the difference of the score between competing models) is generally skewed, and more so
when dealing with small $\nte$ or scoring rules involving high quantiles ($\qs_p$ or $\is_p$ with high $p$).
If the distribution of the (difference of the) scores is skewed,
non-mean statistics such as the mean rank, the win rate or the median score
can pick the wrong model even if the data set contains a huge number of time series.
Since model selection with scoring rules requires estimating the expected value of the score, the mean scaled score is  the most appropriate summary statistic, even if 
its decision might depend on the chosen scaling factor.
Yet in our experiments  the decision of the mean scaled score are robust to the choice of scaling factor, whose choice appears to be less critical than previously assumed. 
We anyway recommend validating the conclusions by checking results across two or more scaling factors.
An open problem is how to check the significance of the differences
between competing models. 
Indeed, in large data sets  the Diebold–Mariano test tends to reject the null hypothesis  even with small  differences in mean scaled scores. 
The problem might be addressed by adopting as Bayesian tests with region of practical equivalence  \citep{kruschke2013bayesian}.

\bibliography{biblio}

@book{bulmer1979principles,
  title={Principles of statistics},
  author={Bulmer, Michael George},
  year={1979},
  publisher={Courier Corporation}
}

@article{makridakis_m5_2022,
	title = {M5 accuracy competition: {Results}, findings, and conclusions},
	volume = {38},
	issn = {01692070},
	shorttitle = {M5 accuracy competition},
	language = {en},
	number = {4},
	urldate = {2026-04-15},
	journal = {International Journal of Forecasting},
	author = {Makridakis, Spyros and Spiliotis, Evangelos and Assimakopoulos, Vassilios},
	month = oct,
	year = {2022},
	pages = {1346--1364},
}

@article{koning_m3_2005,
	title = {The {M3} competition: {Statistical} tests of the results},
	volume = {21},
	copyright = {https://www.elsevier.com/tdm/userlicense/1.0/},
	issn = {01692070},
	shorttitle = {The {M3} competition},
	language = {en},
	number = {3},
	urldate = {2026-04-15},
	journal = {International Journal of Forecasting},
	author = {Koning, Alex J. and Franses, Philip Hans and Hibon, Michèle and Stekler, H.O.},
	month = jul,
	year = {2005},
	pages = {397--409},
}

@article{demvsar2006statistical,
  title={Statistical comparisons of classifiers over multiple data sets},
  author={Dem{\v{s}}ar, Janez},
  journal={{Journal of Machine Learning Research}},
  volume={7},
  number={Jan},
  pages={1--30},
  year={2006}
}

@article{long_scalable_2025,
	title = {Scalable probabilistic forecasting in retail with gradient boosted trees: {A} practitioner’s approach},
	volume = {279},
	issn = {09255273},
	shorttitle = {Scalable probabilistic forecasting in retail with gradient boosted trees},
	language = {en},
	urldate = {2025-03-12},
	journal = {International Journal of Production Economics},
	author = {Long, Xueying and Bui, Quang and Oktavian, Grady and Schmidt, Daniel F. and Bergmeir, Christoph and Godahewa, Rakshitha and Lee, Seong Per and Zhao, Kaifeng and Condylis, Paul},
	month = jan,
	year = {2025},
	pages = {109449},
}

@book{boylan_intermittent_2021,
	title = {Intermittent demand forecasting: context, methods and applications},
	language = {eng},
	publisher = {Wiley},
	author = {Boylan, John E. and Syntetos, Aris A.},
	year = {2021},
}

@article{kolassa_commentary_2022,
	title = {Commentary on the {M5} forecasting competition},
	volume = {38},
	issn = {01692070},
	language = {en},
	number = {4},
	urldate = {2023-05-30},
	journal = {International Journal of Forecasting},
	author = {Kolassa, Stephan},
	month = oct,
	year = {2022},
	pages = {1562--1568},
}

@misc{shchur2025fev,
      title={fev-bench: A Realistic Benchmark for Time Series Forecasting}, 
      author={Oleksandr Shchur and Abdul Fatir Ansari and Caner Turkmen and Lorenzo Stella and Nick Erickson and Pablo Guerron and Michael Bohlke-Schneider and Yuyang Wang},
      year={2026},
      eprint={2509.26468},
      archivePrefix={arXiv},
      primaryClass={cs.LG},
      url={https://arxiv.org/abs/2509.26468}, 
}

@article{kruschke2013bayesian,
  title={Bayesian estimation supersedes the t test.},
  author={Kruschke, John K},
  journal={{Journal of experimental psychology: General}},
  volume={142},
  number={2},
  pages={573},
  year={2013},
  publisher={American Psychological Association}
}

@article{syntetos2012demand,
  title={On the demand distributions of spare parts},
  author={Syntetos, Aris A and Babai, Mohamed Zied and Altay, Nezih},
  journal={International Journal of Production Research},
  volume={50},
  number={8},
  pages={2101--2117},
  year={2012},
  publisher={Taylor \& Francis}
}

@article{hyndman_another_2006,
	title = {Another look at measures of forecast accuracy},
	volume = {22},
	copyright = {https://www.elsevier.com/tdm/userlicense/1.0/},
	issn = {01692070},
	language = {en},
	number = {4},
	urldate = {2025-08-06},
	journal = {International Journal of Forecasting},
	author = {Hyndman, Rob J. and Koehler, Anne B.},
	month = oct,
	year = {2006},
	pages = {679--688},
}

@article{makridakis2022m5,
  title={{The M5 uncertainty competition: results, findings and conclusions}},
  author={Makridakis, Spyros and Spiliotis, Evangelos and Assimakopoulos, Vassilios and Chen, Zhi and Gaba, Anil and Tsetlin, Ilia and Winkler, Robert L},
  journal={International Journal of Forecasting},
  volume={38},
  number={4},
  pages={1365--1385},
  year={2022},
  publisher={Elsevier}
}

@article{svetunkov2023iets,
  title={{iETS: State space model for intermittent demand forecasting}},
  author={Svetunkov, Ivan and Boylan, John E},
  journal={International Journal of Production Economics},
  volume={265},
  pages={109013},
  year={2023},
  publisher={Elsevier}
}

@article{gneiting2007strictly,
	title        = {{Strictly proper scoring rules, prediction, and estimation}},
	author       = {Gneiting, Tilmann and Raftery, Adrian E},
	year         = 2007,
	journal      = {Journal of the American statistical Association},
	publisher    = {Taylor \& Francis},
	volume       = 102,
	number       = 477,
	pages        = {359--378}
}

@article{gneiting2014probabilistic,
  title={Probabilistic forecasting},
  author={Gneiting, Tilmann and Katzfuss, Matthias},
  journal={Annual Review of Statistics and Its Application},
  volume={1},
  number={1},
  pages={125--151},
  year={2014},
  publisher={Annual Reviews}
}

@article{athanasopoulos2023evaluation,
  title={On the evaluation of hierarchical forecasts},
  author={Athanasopoulos, George and Kourentzes, Nikolaos},
  journal={International Journal of Forecasting},
  volume={39},
  number={4},
  pages={1502--1511},
  year={2023},
  publisher={Elsevier}
}

@misc{Tibshirani2023ForecastScoring,
  author       = {Tibshirani, Ryan J.},
  title        = {Forecast Scoring and Calibration},
  howpublished = {Lecture notes},
  institution  = {University of California, Berkeley},
  course       = {Advanced Topics in Statistical Learning},
  year         = {2023},
  url          = {https://www.stat.berkeley.edu/~ryantibs/statlearn-s23/lectures/calibration.pdf}
}

@article{fakoor2023flexible,
  title={Flexible model aggregation for quantile regression},
  author={Fakoor, Rasool and Kim, Taesup and Mueller, Jonas and Smola, Alexander J and Tibshirani, Ryan J},
  journal={Journal of Machine Learning Research},
  volume={24},
  number={162},
  pages={1--45},
  year={2023}
}

@article{spiliotis2021product,
  title={{Product sales probabilistic forecasting: An empirical evaluation using the M5 competition data}},
  author={Spiliotis, Evangelos and Makridakis, Spyros and Kaltsounis, Anastasios and Assimakopoulos, Vassilios},
  journal={International Journal of Production Economics},
  volume={240},
  pages={108237},
  year={2021},
  publisher={Elsevier}
}

@article{hewamalage2023forecast,
  title={Forecast evaluation for data scientists: common pitfalls and best practices},
  author={Hewamalage, Hansika and Ackermann, Klaus and Bergmeir, Christoph},
  journal={Data Mining and Knowledge Discovery},
  volume={37},
  number={2},
  pages={788--832},
  year={2023},
  publisher={Springer}
}

@article{damato2025forecasting,
  title={{Forecasting intermittent time series with Gaussian Processes and Tweedie likelihood}},
  author={Damato, Stefano and Azzimonti, Dario and Corani, Giorgio},
  journal={International Journal of Forecasting},
  year={2026},
  doi = {https://doi.org/10.1016/j.ijforecast.2025.10.001},
  publisher={Elsevier},
  volume  = {in press},
}

@book{Durrett2019,
  author    = {Durrett, Rick},
  title     = {Probability: Theory and Examples},
  publisher = {Cambridge University Press},
  year      = {2019},
  edition   = {5th}
}

@article{bolin_wallin_2023,
title = "Local scale invariance and robustness of proper scoring rules",
author = "David Bolin and Jonas Wallin",
year = "2023",
language = "English",
volume = "38",
pages = "140--159",
journal = "Statistical Science",
issn = "0883-4237",
publisher = "Institute of Mathematical Statistics",
number = "1",
}

@Manual{fableintermittent,
    title = {fable.intermittent: Forecasting Models for Intermittent Time Series},
    author = {Stefano Damato and Lorenzo Zambon and Dario Azzimonti},
    year = {2026},
    note = {R package version 0.1.0},
    url = {https://CRAN.R-project.org/package=fable.intermittent},
    doi = {10.32614/CRAN.package.fable.intermittent},
  }

@article{wheatcroft2019interpreting,
title = {Interpreting the skill score form of forecast performance metrics},
journal = {International Journal of Forecasting},
volume = {35},
number = {2},
pages = {573-579},
year = {2019},
issn = {0169-2070},
author = {Edward Wheatcroft}
}
\appendix

\section{Median and relative scores for the simulation of Sec.\ref{sect:simul_model_selection}}
\label{app:median_scores}

\begin{table}[!h]
\centering
\begin{tabular}{clcccccc}
\toprule
\multicolumn{2}{c}{ } & \multicolumn{2}{c}{\textbf{CRPS}} & \multicolumn{2}{c}{\textbf{QS}$\mathbf{_{0.9}}$} & \multicolumn{2}{c}{\textbf{QS}$\mathbf{_{0.99}}$} \\
 &  & Median score & Rel.~score & Median score & Rel.~score & Median score & Rel.~score\\
\cmidrule(lr){3-4} \cmidrule(lr){5-6} \cmidrule(lr){7-8}
 & \textit{True} & \cellcolor{backcolour}\textcolor{codegreen}{\textbf{0.530}} & 1.000 & 0.316 & 1.000 & 0.047 & 1.000\\
\multirow{-2}{*}{\centering\arraybackslash $\nte = 4$} & \textit{Misspec.} & 0.533 & \cellcolor{backcolour}\textcolor{red}{\textbf{0.995}} & \cellcolor{backcolour}\textcolor{red}{\textbf{0.305}} & \cellcolor{backcolour}\textcolor{red}{\textbf{0.988}} & \cellcolor{backcolour}\textcolor{red}{\textbf{0.042}} & \cellcolor{backcolour}\textcolor{red}{\textbf{0.934}}\\
\addlinespace[0.5em]
 & \textit{True} & \cellcolor{backcolour}\textcolor{codegreen}{\textbf{0.560}} & \cellcolor{backcolour}\textcolor{codegreen}{\textbf{1.000}} & \cellcolor{backcolour}\textcolor{codegreen}{\textbf{0.342}} & \cellcolor{backcolour}\textcolor{codegreen}{\textbf{1.000}} & 0.048 & \cellcolor{backcolour}\textcolor{codegreen}{\textbf{1.000}}\\
\multirow{-2}{*}{\centering\arraybackslash $\nte = 28$} & \textit{Misspec.} & 0.563 & 1.005 & 0.348 & 1.014 & \cellcolor{backcolour}\textcolor{red}{\textbf{0.045}} & 1.024\\
\addlinespace[0.5em]
 & \textit{True} & \cellcolor{backcolour}\textcolor{codegreen}{\textbf{0.563}} & \cellcolor{backcolour}\textcolor{codegreen}{\textbf{1.000}} & \cellcolor{backcolour}\textcolor{codegreen}{\textbf{0.349}} & \cellcolor{backcolour}\textcolor{codegreen}{\textbf{1.000}} & \cellcolor{backcolour}\textcolor{codegreen}{\textbf{0.050}} & \cellcolor{backcolour}\textcolor{codegreen}{\textbf{1.000}}\\
\multirow{-2}{*}{\centering\arraybackslash $\nte = 100$} & \textit{Misspec.} & 0.567 & 1.006 & 0.356 & 1.018 & 0.054 & 1.061\\
\bottomrule
\end{tabular}

\caption{
Model selection over $10^5$ simulations.
For the selected models we highlight the background;
we use green when the true model is selected, and red otherwise.\\
For short test sets and/or high quantiles, both the median score and the relative score select the misspecified model. Similar to the mean rank, on $\qs_{0.99}$ even for $\nte = 28$ the wrong model is selected. On the contrary, the true distribution has a better CRPS for any test set size. Indeed, eq.~\eqref{eq:crps} shows that it takes into account all quantile levels, balancing the effect of the $\qs_p$ where underforecasts are rare yet strongly penalised.
}
\label{tab:median_rel_score}
\end{table}

\newpage
\section{Results on  Auto  data sets}\label{appendix:auto}
\begin{figure}[!ht]
    \centering
    \includegraphics[width=1\linewidth]{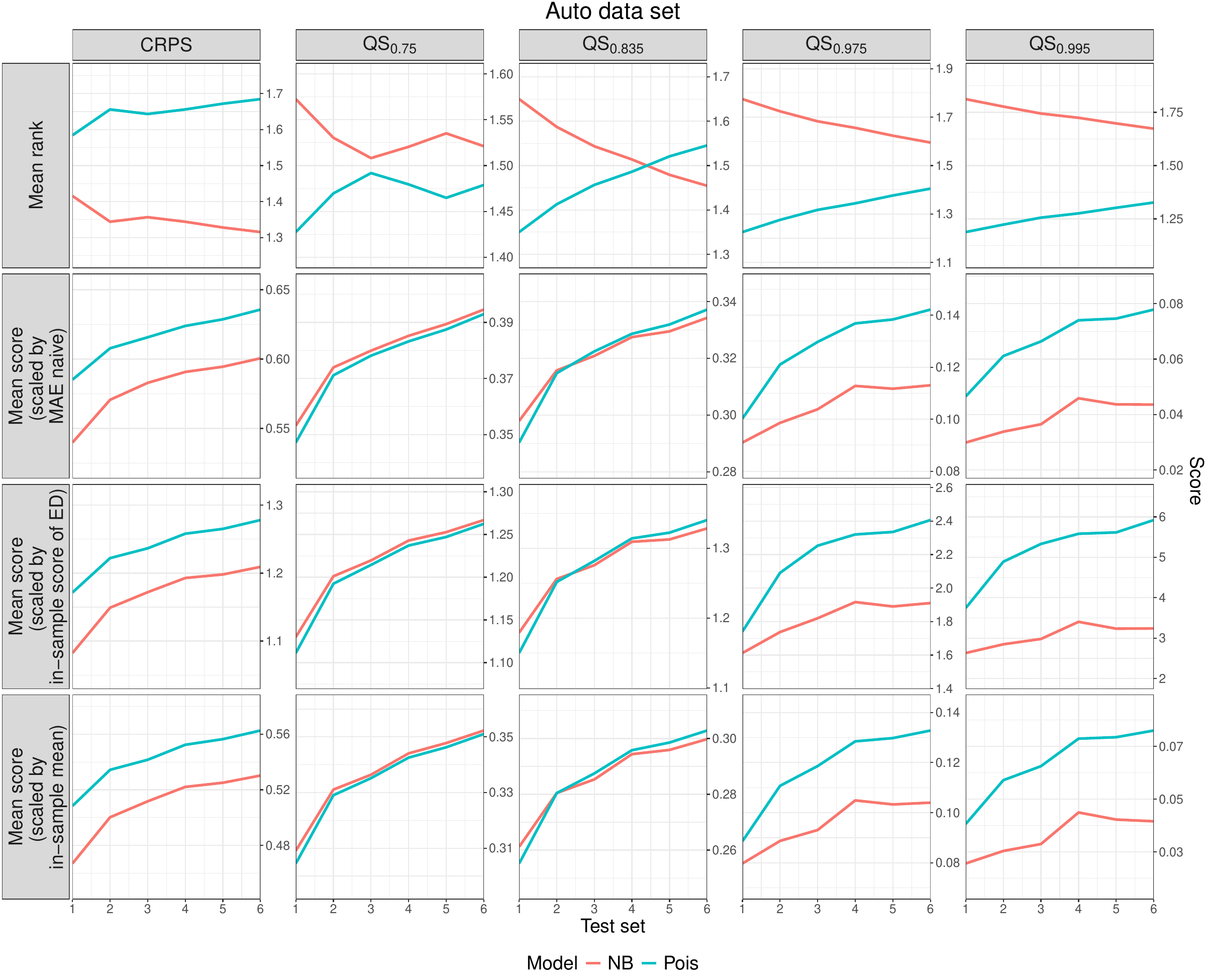}
    \caption{Mean ranks and mean scaled scores on the Auto data set. 
    }
    \label{fig:Auto_experiment_full}
\end{figure}

\newpage
\section{Results on  RAF  data set}\label{appendix:raf}
\begin{figure}[!ht]
    \centering
    \includegraphics[width=1\linewidth]{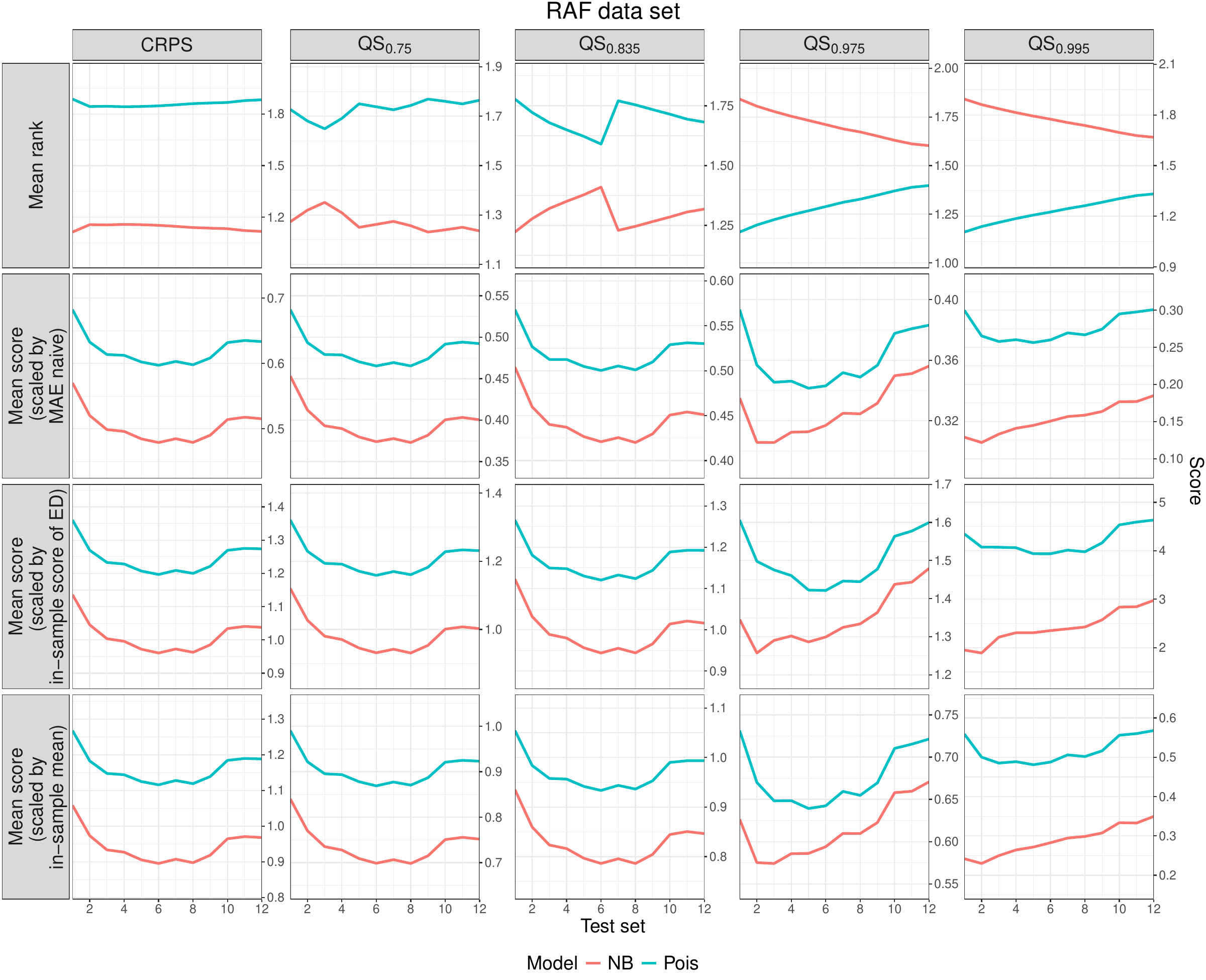}
    \caption{Mean ranks and mean scaled scores on the RAF data set.}
    \label{fig:RAF_experiment_full}
\end{figure}

\newpage
\section{Full results on  M5 data set}\label{appendix:m5}

\begin{figure}[!ht]
    \centering
    \includegraphics[width=1\linewidth]{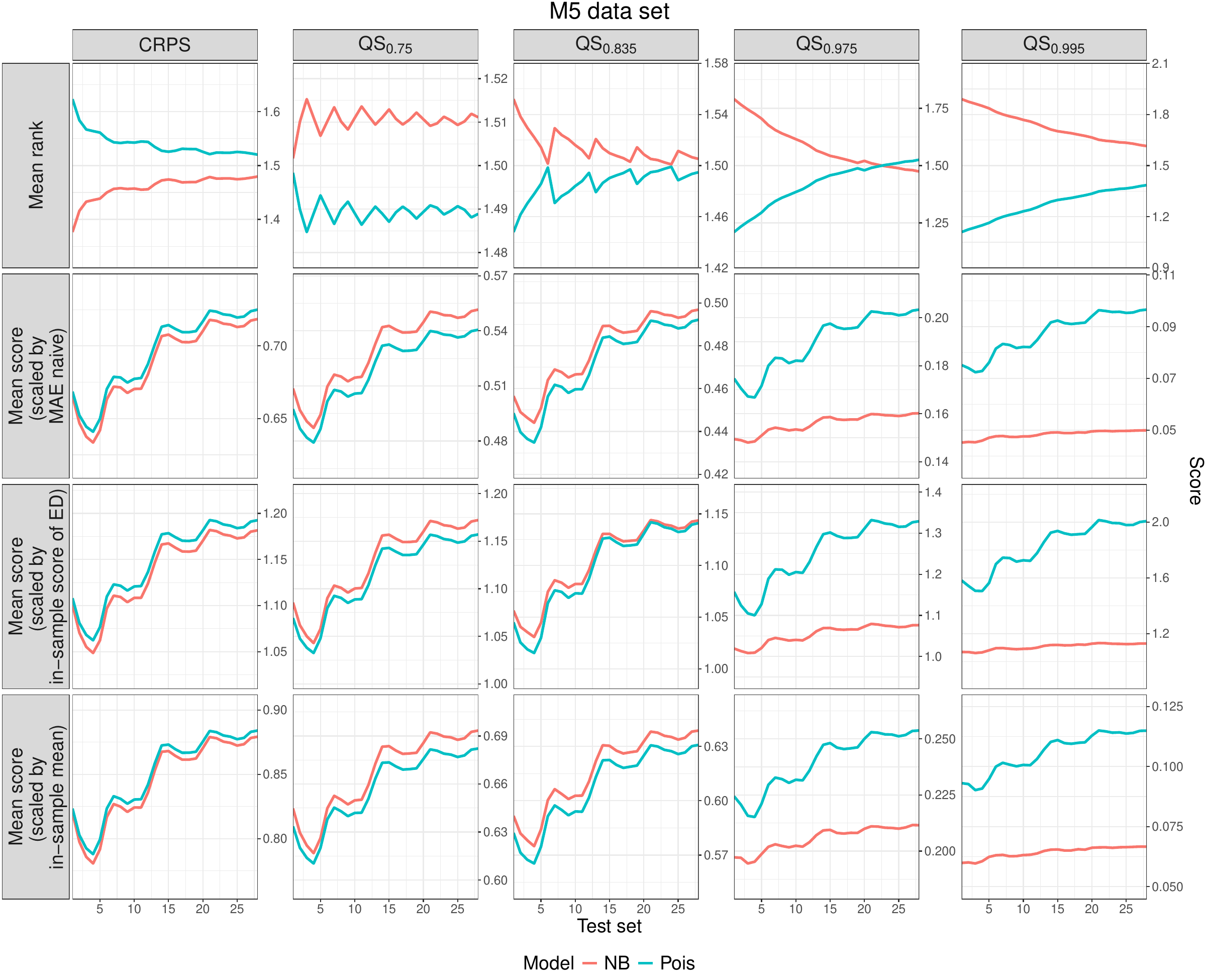}
    \caption{Mean ranks and mean scaled scores on the M5 data set. 
    }
    \label{fig:M5_experiment_full}
\end{figure}

\end{document}